\documentclass[twoside,11pt]{article}
\usepackage{xcolor}
\usepackage[colorlinks=true,
            linkcolor=blue,
            citecolor=blue,
            urlcolor=blue]{hyperref}

\usepackage{blindtext}
\usepackage{svg}
\usepackage{graphicx}
\usepackage{amsmath}
\usepackage{booktabs}
\usepackage[linesnumbered,ruled,vlined]{algorithm2e}

\usepackage{dmlr2e}

\usepackage{lastpage}
\dmlrheading{25}{2026}{1-\pageref{LastPage}}{1/26}{8/26}{164}{Syed Mumtahin Mahmud, Mahdi Mohd Hossain Noki, Prothito Shovon Majumder, Abdul Mohaimen Al Radi, Sudipto Das Sukanto, Afia Lubaina, and Md. Mosaddek Khan} 
\def\openreview{\url{https://openreview.net/forum?id=38E3mAI0G4}} 

\ShortHeadings{SloMoDeblur: A Large-Scale Smartphone Image Deblurring Dataset}{Mahmud et al.}
\firstpageno{1}
\hypersetup{
    colorlinks=true,
    linkcolor=blue,
    citecolor=blue,
    urlcolor=blue
}

\begin{document}

\title{SloMoDeblur: A Large-Scale Smartphone Image Deblurring Dataset}

\author{
\name Syed Mumtahin Mahmud \email syedmumtahin-2019617833@cs.du.ac.bd
\AND
\name Mahdi Mohd Hossain Noki \email mahdimohdhossain-2019017785@cs.du.ac.bd
\AND
\name Prothito Shovon Majumder \email prothitoshovon-2018725302@cs.du.ac.bd
\AND
\name Abdul Mohaimen Al Radi \email abdulmohaimenal-2018925300@cs.du.ac.bd
\AND
\name Sudipto Das Sukanto \email sudiptodas-2020215639@cs.du.ac.bd
\AND
\name Afia Lubaina \email afia-2020815624@cs.du.ac.bd
\AND
\name Md.~Mosaddek Khan \email mosaddek@du.ac.bd
\\ \\
\addr
Cognitive Agents and Interaction Lab, Department of Computer Science and Engineering, University of Dhaka, Dhaka-1000, Bangladesh
}

\editor{Fernando Perez-Cruz}

\maketitle

\begin{abstract}
Motion blur remains one of the most common and visually disruptive degradations in real-world smartphone imaging, yet existing deblurring benchmarks are often limited in scale, resolution, or domain relevance. This gap is especially pronounced for smartphones, where rolling shutter, small sensors, and ISP processing produce blur statistics that differ from GoPro/DSLR-based benchmarks. We introduce a large-scale smartphone-oriented deblurring dataset constructed from 240~fps slow-motion video. To approximate exposure-time radiance integration, we synthesize blur by temporally averaging a fixed window of $N=30$ consecutive frames, which corresponds to an effective exposure of $T=1/8$~second, and we select the temporally centered frame as the sharp ground truth. The resulting benchmark contains 42,045 paired blur--sharp images at $1920\times1080$ resolution spanning 843 distinct scenes, with a train/test split of 37,841/4,204 pairs. We benchmark multiple state-of-the-art deblurring models using PSNR and SSIM and observe consistent performance degradation relative to the baseline similarity between the input blurry images and ground truth, underscoring the realism and difficulty of the proposed data. We release the dataset and generation scripts via HuggingFace to facilitate the development and evaluation of robust, deployment-oriented deblurring methods. The dataset is available at \texttt{\url{https://huggingface.co/datasets/masterda/SloMoBlur}}.
\end{abstract}

\begin{keywords}
  Image deblurring, Smartphone imaging, High-speed video datasets
\end{keywords}

\section{Introduction}
\label{section:introduction}
\begin{figure}[ht]
    \centering
    \includegraphics[width=\linewidth]{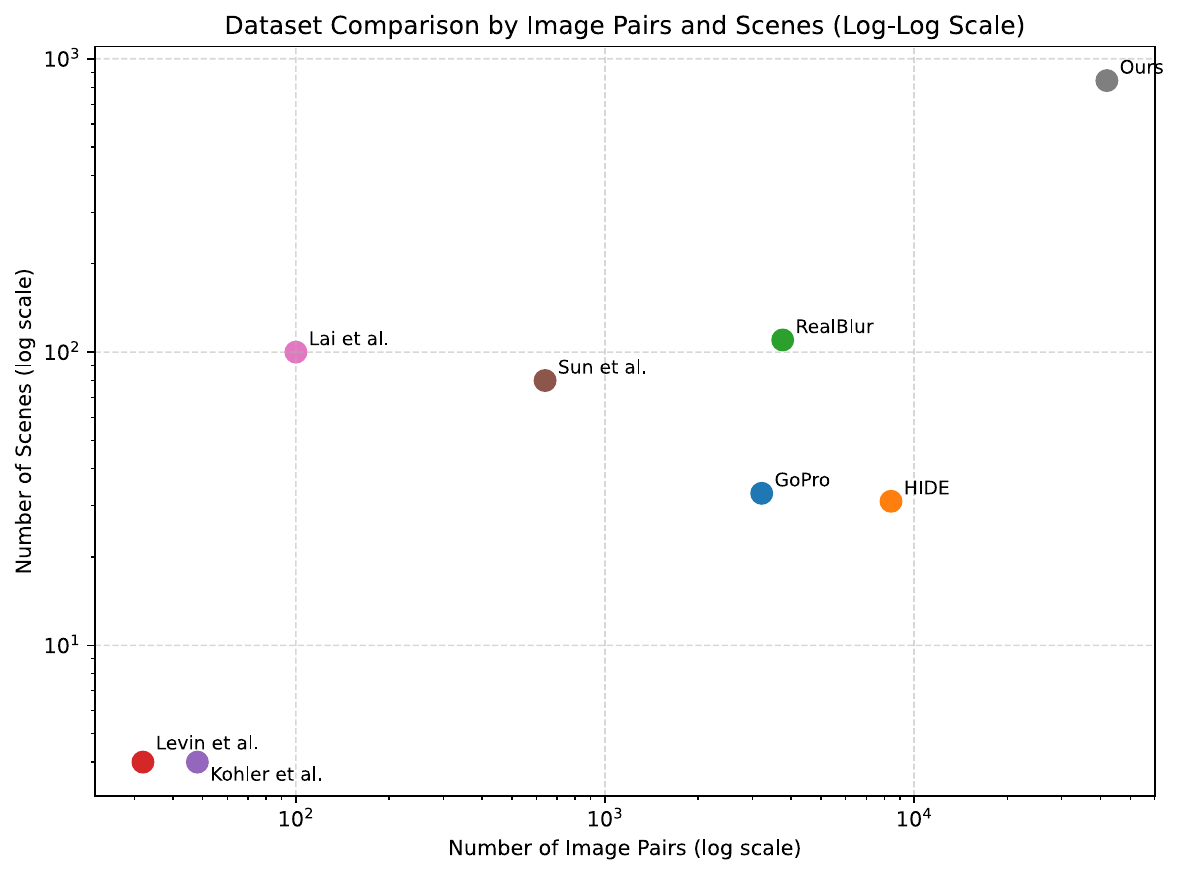}
    \caption{A comparison with widely used deblurring datasets reveals that SloMoDeblur offers significantly higher scale and broader scene coverage, occupying the top-right quadrant distinct from previous datasets.}
    \label{fig:dataset-comparison}
\end{figure}

Image restoration is a widely studied area in computer vision, with extensive research devoted to improving image quality by mitigating degradations such as noise, haze, moiréing, and blur. Among these forms of degradation, motion blur remains one of the most common and visually disruptive artifacts. Image deblurring refers to the process of estimating a latent sharp image from a blurred observation. The formation of motion blur is influenced by several factors, including fast-moving subjects, camera shake, occlusions along motion boundaries, and variations in scene depth. These factors produce blur patterns that are often spatially varying and difficult to characterize analytically.

Early approaches to image deblurring \cite{gupta, whyte, harmeling, hirsch, cho, pandcp, fergus, qishan, xu1, xu2} relied on simplified blur models under the assumption of uniform or locally linear blur. These methods typically adopt the following spatially invariant formulation:
\[
B = I \otimes k + \eta ,
\]
where $I$, $B$, and $k$ denote the sharp image, blurred image, and blur kernel, respectively. The operator $\otimes$ denotes convolution and $\eta$ refers to additive noise. Although analytically convenient, this model rarely captures the complexity of real-world motion blur, which is often non-uniform and spatially heterogeneous. Consequently, jointly estimating $I$ and $k$ becomes highly ill-posed when applied to natural images from diverse environments.

The introduction of data-driven learning methods changed the landscape of image deblurring. Convolutional neural network-based approaches were among the earliest learning-based methods, often trained on synthetically generated blur using uniform kernels \cite{ schuler, xuli, chakra}, or designed to estimate locally linear blur kernels \cite{sunjian}. However, these methods were constrained by the lack of large-scale datasets containing aligned blurred, and sharp image pairs. More recently, transformer-based and state-space model-based approaches have shown significant progress but remain heavily dependent on the datasets used during training.

Most existing deblurring datasets, such as GoPro and HIDE, are captured using high-speed GoPro cameras, while datasets such as RealBlur rely on dual-exposure DSLR setups \cite{Nah_2017_CVPR, rim_2020_ECCV}. These devices use large sensors, dedicated optics, and strong hardware-level stabilization. As a result, the captured blur patterns often differ significantly from those produced by smartphones. Modern smartphones, which are now the most widely used imaging devices in daily life, operate with small sensors and heavily rely on computational photography. This leads to blur that is more severe, more complex, and strongly influenced by rolling shutter distortion, and aggressive in-camera processing. Models trained primarily on DSLR or GoPro-based datasets may experience performance degradation when applied to smartphone images, motivating the development of smartphone-oriented benchmarks.

Constructing a realistic blur-sharp dataset is challenging. The most fundamental challenge is to ensure precise geometric and photometric alignment between blurred frames and their corresponding sharp references. Since camera motion is typically required to generate blur, capturing perfectly aligned pairs is difficult. In addition, the dataset must adequately represent diverse real-world scenarios where smartphone motion blur commonly occurs, while maintaining minimal noise in the ground truth sharp frames.

To address these challenges, we adopt the high-frame-rate temporal integration approach used in GoPro and HIDE. Motion blur can be viewed as the temporal integral of scene radiance over the exposure period, and averaging densely sampled high-speed frames provides a practical discrete approximation of this physical process. This strategy captures real smartphone motion trajectories and micro-motion patterns, and is consistent with rolling-shutter characteristics present in smartphone imaging. Therefore, it produces blur that better reflects smartphone capture conditions than DSLR-based long-exposure blur.

Using the 240 fps slow-motion recording capability of modern iPhones, we generate long exposure blur images by averaging a sequence of consecutive frames and selecting the middle frame as the sharp ground truth. This approach ensures alignment while closely approximating real-world exposure-time blur. Following this procedure, we construct a dataset of 42,000 blurred and sharp image pairs, significantly larger than existing benchmarks such as GoPro and RealBlur, which contain only a few thousand samples. Our dataset spans diverse indoor and outdoor scenes and is specifically tailored to the imaging characteristics of contemporary smartphones. In summary, our work makes the following contributions:

\begin{itemize}
    \item We introduce a large-scale smartphone-based motion deblurring dataset consisting of 42,000 blur-sharp image pairs, created using high frame rate temporal integration to ensure realistic motion blur and precise alignment.
    \item We address the domain gap present in existing DSLR and GoPro-oriented datasets by focusing specifically on smartphone imagery, which better reflects real-world user photography and everyday capture conditions.
    \item We benchmark several state-of-the-art deblurring models on our smartphone dataset and observe substantial performance drops compared to their results on existing benchmarks, highlighting the increased complexity of smartphone blur and motivating the need for more generalizable deblurring approaches.
\end{itemize}

\section{Related Works}

In this section, we review existing image deblurring models and datasets, and we highlight the limitations that motivate the need for a large-scale smartphone-specific deblurring benchmark.

\subsection{Image Deblurring Models}

The image deblurring problem has historically been approached in two distinct ways. The first is blind image deblurring, where the goal is to estimate both the latent sharp image and the unknown blur kernel. Classical blind deblurring methods rely on carefully designed priors or optimization schemes to handle uniform or spatially-varying blur, such as methods from \cite{gupta, whyte, harmeling, hirsch, fergus, xu2, schuler, xu1, qishan, pandcp, cho}. These methods provided early insights into the structure of blur but often struggled with real-world scenes due to the complexity and irregularity of natural motion.

The second direction focuses on direct restoration using deep learning, where a model is trained to map blurry images to their sharp counterparts. With the rise of powerful architectures and the availability of larger datasets, learning-based methods have achieved substantial improvements. Recent state-of-the-art models routinely report PSNR values above 34 dB and SSIM scores approaching or exceeding 0.97. Many of these advancements are driven by innovations in network design, such as adaptive reversible decoders \cite{adarevd}, efficient state-space representations \cite{kong2024efficientvisualstatespace}, and a growing reliance on transformer-based architectures \cite{Kong_2023_CVPR, xintm2024LoFormer, Zamir2021Restormer}, which capture long-range dependencies more effectively than traditional CNNs \cite{dmphn, mprnet, mimounet}.

In parallel, frequency-domain processing has emerged as a complementary strategy. Models such as DeepRFT+ \cite{deeprft} and FSNet \cite{fsnet} incorporate frequency-aware components to exploit the spectral characteristics of motion blur. \cite{icaart26} further explored a hybrid spatial-frequency approach by integrating Vision Transformer-based feature modeling with FFT-ReLU-based frequency processing for image deblurring. Lightweight networks, test-time refinement strategies, and Mamba-based state-space models are increasingly being explored to improve generalization and efficiency.

Despite these advances, modern deblurring systems still struggle in real-world dynamic scenes where blur is severe, heterogeneous, and influenced by device-specific imaging pipelines. As discussed in the introduction, most existing models are trained on DSLR or GoPro-style datasets, and therefore exhibit substantial performance degradation when applied to smartphone images, which now constitute the primary source of real-world photography.

\subsection{Synthetic Datasets}

Synthetic datasets have historically served as the foundation for benchmarking and early learning-based deblurring. \cite{LevinEval} introduced a controlled dataset of 32 images blurred using real camera shake, while \cite{SunDataset} proposed synthetically blurred images generated by convolving 80 sharp images with a small set of predefined kernels. \cite{KohlerDataset} extended this direction by recording actual camera motion trajectories to simulate spatially varying blur. Although these datasets were instrumental for evaluating kernel estimation methods, their limited scale makes them unsuitable for training modern deep neural networks.

The introduction of the GoPro dataset \cite{Nah_2017_CVPR} marked a significant shift toward large-scale synthetic data for learning-based deblurring. Blur was simulated by averaging high-frame-rate video sequences captured using a GoPro action camera. This method was later adapted for human-centric scenes in the HIDE dataset \cite{HIDE}, which focuses on people in motion but follows a similar data generation strategy.

While these datasets have fueled notable progress, they inherit the imaging characteristics of GoPro devices, which differ substantially from smartphones in optics, sensor behavior, noise patterns, and stabilization. As a result, models trained on these datasets often fail to generalize to the more complex and noisier motion blur produced by modern mobile cameras.

\subsection{Datasets Captured Under Natural Imaging Conditions}

Real-world datasets attempt to capture naturally occurring blur without relying on synthetic averaging. Early efforts, such as those by \cite{LevinEval} and \cite{KohlerDataset}, introduced real blur under controlled motion; however, their small scale and lack of paired sharp images limited their utility for supervised learning. \cite{LaiDataset} later proposed a dataset of 100 real blurry images, but the absence of aligned sharp references constrained applicability to evaluation only.

The RealBlur dataset \cite{rim_2020_ECCV} addressed several of these shortcomings by using a beam-splitter setup to capture long-exposure blur and short-exposure sharp images simultaneously. This design produces accurately aligned blur–sharp pairs under natural lighting and is one of the first real-world datasets suitable for both training and evaluation. However, RealBlur remains limited in scale, with fewer than 3,758 image pairs, and its beam-splitter setup reduces sensor illumination, requiring significant post-processing. Moreover, because it was captured using DSLR cameras, its content and imaging characteristics do not reflect the behavior of consumer smartphones, which dominate everyday photography.

\subsection{Challenges}

Despite the influence of both synthetic and real-world datasets, several critical gaps remain. Large-scale synthetic datasets such as GoPro and HIDE contain only a few thousand samples, insufficient for training the increasingly large and complex architectures used in state-of-the-art deblurring. Their focus on DSLR-style or GoPro-style imaging results in a noticeable domain shift that limits real-world generalization. Although RealBlur improves realism and introduces true blur–sharp alignment, it is limited in scale and biased toward high-quality DSLR imagery. Smartphone cameras, with their smaller sensors, computational imaging pipelines, higher noise levels, and susceptibility to severe motion blur, produce fundamentally different degradations. Yet, the field currently lacks a large-scale, smartphone-specific dataset that adequately reflects the real-world conditions under which most users capture images. This gap motivates the need for a new dataset designed specifically around smartphone imaging characteristics, addressing the shortcomings of existing resources and enabling more robust, generalizable deblurring models.

\section{Dataset Construction}

 Our objective is to construct a large-scale, smartphone-specific dataset that faithfully represents the motion blur characteristics encountered in everyday photography. Existing DSLR and GoPro-based datasets provide valuable insights, but their imaging pipelines, sensor sizes, and stabilization mechanisms differ significantly from those of modern smartphones. To bridge this domain gap, we design a data acquisition pipeline that captures sharp and blurred pairs using the native hardware and computational imaging stack found in real consumer devices. This approach ensures that the resulting dataset reflects both the physical constraints and the algorithmic behaviors inherent to contemporary mobile cameras.

\subsection{Acquisition Hardware and Rationale}

All data in our dataset is captured using the iPhone 15 Pro, a widely available consumer-grade device featuring a high-quality sensor, robust dynamic range, and hardware-supported slow-motion recording. This choice serves two purposes. First, the 240 fps high-frame-rate mode allows dense temporal sampling of real-world motion, enabling accurate approximation of the radiance integration that causes motion blur. Second, it ensures the dataset aligns naturally with the conditions and devices used by the majority of end-users. Unlike datasets collected using DSLR or GoPro systems, our data is consistent with the imaging characteristics of smartphone cameras, including smaller sensor sizes, limited light-capture capacity, rolling-shutter readout characteristics of CMOS sensors \cite{CMOS}, and subtle hand-induced motion trajectories.

To maintain benchmark stability and reproducibility, we adopt a single-device capture protocol rather than aggregating across heterogeneous smartphone models. This design mirrors the principles adopted by established benchmarks such as GoPro and RealBlur, where consistency in imaging characteristics is prioritized over device diversity. While modern smartphones from different manufacturers (e.g., Samsung, Pixel, Xiaomi) often use Sony IMX-series sensors with comparable raw capture properties, the iPhone’s high-speed video pipeline provides superior frame-timing stability and reliable hardware-level stabilization. These characteristics are essential for constructing blur–sharp pairs with minimal geometric misalignment during temporal averaging. Thus, SloMoDeblur should be understood as a controlled single-device smartphone benchmark, rather than a cross-device evaluation of all smartphone imaging pipelines.

\subsection{High-Speed Video Capture}

We record all video sequences at 240 frames per second with a native 1920\,$\times$\,1080 spatial resolution. This configuration provides the temporal granularity needed to approximate the continuous exposure process while preserving sufficient spatial resolution for evaluating fine textures and edges. Using a widely available mobile device rather than a specialized camera platform lowers the barrier for future dataset extensions and facilitates reproducibility across research groups.

The high frame rate is particularly important because it minimizes inter-frame motion, ensuring that the center frame within a short temporal window can serve as a reliable approximation of a sharp image. Simultaneously, accumulating many consecutive frames enables the construction of realistic blur with naturally occurring motion trajectories and rolling-shutter distortions characteristic of smartphone operation.

\subsection{Blur Generation Strategy}

Motion blur in a physical camera is formed through the temporal integration of scene radiance during the exposure interval. To approximate this process, we simulate blur by averaging multiple consecutive frames from the 240 fps video stream. While straightforward, this approach remains physically grounded and scalable to thousands of scenes.

We empirically investigated several exposure durations to model realistic smartphone blur. Long exposures approaching one second produced undesirable nonlinearities, including saturation, aggressive noise amplification, and severe detail loss. We observed that averaging approximately 30 frames at 240 fps (equivalent to an exposure time of $1/8$ second) offered an effective compromise: it introduced complex yet plausible motion blur while preserving photometric fidelity. This choice maintains dynamic range, prevents signal clipping, and captures typical hand-held motion patterns without introducing extreme artifacts not representative of real-world use.

Although averaging assumes approximately uniform motion within the window, the range of motions present in smartphone usage makes this a practical and widely adopted approximation. Furthermore, our collection pipeline ensures consistent application of this methodology across all captured sequences, creating a uniform basis for training and evaluation.

\begin{figure}[ht]
    \centering
    \includegraphics[width=\linewidth]{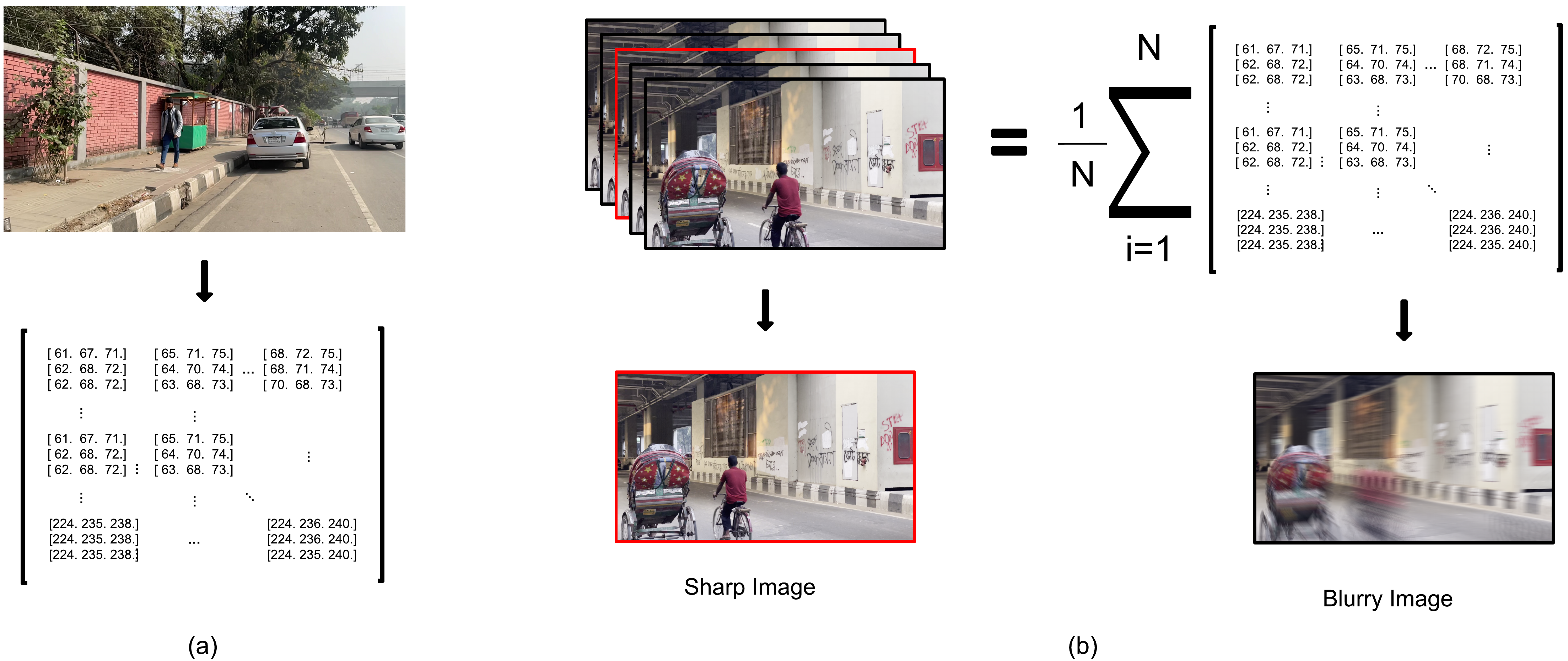}
    \caption{(a) Extracting pixel data from high-speed video frames. (b) Layering temporally adjacent frames to generate synthetic smartphone motion blur.}
    \label{fig:SLoMoDeblurNeurips}
\end{figure}

\subsection{Blur Formation Model}

Motion blur is defined as the temporal integral of sharp frames over exposure duration $T$. Mathematically, the blur image $B$ is modeled as:
\begin{equation}
B = \frac{1}{T} \int_{0}^{T} F(t)\, dt.
\end{equation}
Digital cameras approximate this integral through sampling. We discretize the process as:
\begin{equation}
B^{\prime} = \frac{1}{N} \sum_{i=0}^{N-1} F_i,
\end{equation}
where $F_i$ denotes the $i$-th sharp frame captured at 240 fps. Fixing $N = 30$ produces a blur corresponding to an effective $T = 1/8$ second, aligning well with common smartphone exposures under natural motion.

At 240 fps, the temporal interval between consecutive frames is sufficiently small that inter-frame displacement remains limited for typical handheld smartphone motion. Therefore, averaging 30 consecutive frames provides a practical discrete approximation of the continuous exposure integration process. However, in extreme cases involving very fast nearby object motion, minor discretization artifacts or step-like motion patterns may appear. A direct comparison with true optical long-exposure captures is beyond the scope of this work and remains an important direction for future validation.

The full blur–sharp pair generation procedure is summarized in Algorithm~\ref{alg:composite}. For each non-overlapping sequence of 30 frames, we compute the blurred image by averaging and designate the center frame as the sharp ground truth. 

\begin{algorithm}
\caption{Create Composite Images from Video}
\label{alg:composite}
\KwIn{Video $V$ with frames $V[0 \ldots N-1]$}
\KwOut{Blurred and ground truth image pairs saved to disk}

$FRAMES\_TO\_LAYER \gets 30$\;
$frame\_itr \gets 0$\;

\While{$frame\_itr + FRAMES\_TO\_LAYER \leq N$}{ \label{line:while_start}
    $accumulator \gets 0$ (empty frame)\;
    
    \For{$i \gets 0$ \KwTo $FRAMES\_TO\_LAYER - 1$}{ \label{line:for_start}
        $frame \gets \text{float}(V[frame\_itr + i])$\;
        $accumulator \gets accumulator + frame$\;
        
        \If{$i = \lfloor FRAMES\_TO\_LAYER / 2 \rfloor$}{ \label{line:mid_frame}
            $ground\_truth \gets V[frame\_itr + i]$\;
        }
    } \label{line:for_end}

    $blurred \gets \text{uint8}(accumulator / FRAMES\_TO\_LAYER)$\;
    
    Save $blurred$ to \texttt{images/blurred}\;
    
    Save $ground\_truth$ to \texttt{images/groundtruth}\;
    
    $frame\_itr \gets frame\_itr + FRAMES\_TO\_LAYER$\;
} \label{line:while_end}
\end{algorithm}

We iterate through the video frames (ref~\ref{line:while_start} in non-overlapping segments of 30 frames (line~\ref{line:while_end}). In each iteration, a temporary variable \texttt{accumulator} is initialized to hold the sum of pixel values (line~\ref{line:for_start}). The midpoint frame is selected as the sharp ground truth (line~\ref{line:mid_frame}). Once all 30 frames have been accumulated, their mean is computed and converted to construct the blurred image (line~\ref{line:for_end}).

\subsection{Ground Truth Definition}
\label{subsection:ground-truth-definition}

The midpoint frame within each 30-frame segment is selected as the ground truth sharp reference:
\begin{equation}
F_{\text{GT}} = F_{\lfloor N/2 \rfloor}.
\end{equation}
At 240 fps, the inter-frame movement is sufficiently small that the center frame exhibits negligible motion distortion. This practice follows established methodologies from prior high-frame-rate datasets while being intrinsically aligned with smartphone image formation.

\subsection{Camera Sensor Characteristics}

Smartphones differ fundamentally from DSLRs and GoPro-style devices. Their smaller sensors, narrower apertures, higher reliance on ISP computational enhancements, and susceptibility to rolling-shutter effects introduce blur patterns distinct from those observed in professional camera systems. These characteristics frequently amplify motion blur in low-light or high-motion scenarios, making them both a challenge for existing models and a primary motivation for constructing a smartphone-oriented dataset. Leveraging such characteristics is intended to improve the relevance of learned models for real-world user environments rather than curated professional setups.

\subsection{Dataset Diversity and Real-World Relevance}

We capture a diverse range of scenes, motions, and lighting conditions to reflect the variety found in typical smartphone usage. The dataset includes human-centric activities (walking, gesturing, interacting with objects), dynamic object motions (flying papers, swinging bags), and camera-induced trajectories (panning, shaking, rotational movements). These sequences include both global and local motion, as well as parallax-induced non-uniform blur, depth variation, and multi-object occlusion.

Because the dataset is captured using a consumer device, it naturally incorporates the tone-mapping, stabilization, noise characteristics, and dynamic-range limitations typical of smartphone imaging. All image pairs maintain a stable resolution of 1920\,$\times$\,1080, and pixel values are maintained in a consistent RGB color space extracted directly from the video stream. By prioritizing well-lit scenes during capture, we preserve temporal fidelity during averaging and avoid confounding factors such as elevated noise or ISP-induced artifacts.

Overall, the dataset provides a high-fidelity, diverse, and practically relevant benchmark for learning-based motion deblurring systems. Its grounding in real-world smartphone imaging makes it well suited for developing models that generalize beyond laboratory settings to the everyday scenarios encountered by end users.

\section{Experiments and Evaluation}
This chapter presents a comprehensive evaluation of state-of-the-art image deblurring models on our proposed dataset. We begin by benchmarking several leading methods using standard quantitative metrics such as PSNR and SSIM on a substantial 10\% test split, which includes approximately 4,200 high-resolution image pairs. The performance results are analyzed in the context of a baseline derived from direct blurry-to-ground-truth comparisons, highlighting the inherent difficulty of deblurring images from our dataset. In addition, we provide a comparative overview of existing deblurring datasets, emphasizing how our contribution stands out in terms of scale, diversity, and realism. Finally, we include qualitative visual results that illustrate model behavior across challenging blur conditions, offering deeper insights into the strengths and limitations of current approaches.
\subsection{Benchmarks}
We report the performance of leading deblurring models on our dataset using PSNR and SSIM. This analysis focuses on how well each model handles the realistic and diverse blur patterns present in our test split. In addition to the quantitative results, we include a comparative overview of existing deblurring datasets to highlight how our benchmark differs in scale, complexity, and practical relevance.

\begin{table}[h]
\centering
\caption{Quantitative performance of state-of-the-art deblurring models on our dataset, reported in terms of PSNR and SSIM. The “Blurry vs. GT” row indicates the baseline quality of the input blurry images compared to the ground truth. }
\label{tab:benchmark}
\begin{tabular}{lcc}
\toprule
Model & PSNR (dB) & SSIM \\
\midrule
\textbf{Blurry vs. GT (baseline)} & \textbf{32.38} & \textbf{0.777} \\
Restormer \cite{Zamir2021Restormer} & 31.89 & 0.776 \\
MPRNet \cite{mprnet} & 31.98 & 0.785 \\
NAFNet \cite{chen2022simple} & 31.55 & 0.715 \\
FFTformer \cite{Kong_2023_CVPR} & 32.25 & 0.778 \\
DMPHN \cite{dmphn} & 31.11 & 0.697 \\
MIMO-UNet++ \cite{mimounet} & 31.35 & 0.704 \\
AdaRevD ~\cite{adarevd} & 31.46 & 0.75 \\
\bottomrule
\end{tabular}
\end{table}
Results shown in Table~\ref{tab:benchmark} show that all evaluated models perform below this baseline in PSNR, underscoring the difficulty and realism of our benchmark. The average PSNR and SSIM calculated from the input blurry image and the ground truth are 32.38 and 0.777, which sets the baseline performance. None of the papers crossed the baseline PSNR, but FFTFormer and MPRNet exceeded the baseline SSIM.

One question might be why the \textit{Blurry vs. GT} comparison already yields a relatively high baseline of $32.38$\,dB PSNR and $0.777$ SSIM. This behavior is largely a consequence of how blur--sharp pairs are constructed from high-frame-rate video. Specifically, motion blur is synthesized by temporally averaging $N=30$ consecutive frames captured at $240$\,fps, which discretizes the exposure-time radiance integration model. For each non-overlapping 30-frame segment, the temporally centered (midpoint) frame is selected as the sharp ground truth. At $240$\,fps, inter-frame motion is small enough that the center frame exhibits negligible motion distortion and serves as a reliable sharp reference, while the averaging operation primarily suppresses fine details in regions undergoing motion rather than changing the overall scene layout.  Moreover, because the dataset contains both global and local motion patterns, many pixels can remain approximately stable within a short 30-frame window, so substantial portions of the blurry image remain close to the selected ground truth, inflating pixel-wise similarity and structural agreement.  Finally, by prioritizing well-lit scenes during capture, the averaging procedure preserves temporal and photometric fidelity and avoids elevated noise or ISP-induced artifacts that would otherwise reduce similarity between the blurred observation and the sharp reference. 

\begin{table*}
\centering
\caption{Comparison of widely used image deblurring datasets, highlighting dataset scale, train/test splits, scene diversity, resolution, and data acquisition type. Our dataset provides the largest scale, highest resolution, and broadest scene coverage among existing benchmarks.}
\label{tab:deblurring_datasets}
\normalsize
\setlength{\tabcolsep}{4pt}
\begin{tabular}{lccccc}
\toprule
\textbf{Dataset} & \textbf{Total Pairs} & \textbf{Train/Test} & \textbf{Scenes} & \textbf{Res.} & \textbf{Type} \\
\midrule
GoPro & 3,214 & 2,103 / 1,111 & 33 & $1280{\times}720$ & Synth. \\
HIDE &  8,422 & 6,397 / 2,025 & 31 & $1280{\times}720$ & Synth. (human) \\
RealBlur & 3,758 & 3,058 / 700 & 110 & $1280{\times}720$ & Real \\
Levin et al.  & 32 & – & 4 & $255{\times}255$ & Real (ctrl.) \\
Köhler et al. & 48 & 38 / 10 & 4 & $800{\times}800$ & Real (ctrl.) \\
Sun et al. & 640 & – & 80 & $640{\times}480$ & Synth. \\
Lai et al. & 100 blurry & – & 100 & $1600{\times}1088$ & Real (unpaired) \\
\textbf{Ours} & \textbf{42,045} & \textbf{37,841 / 4,204} & \textbf{843} & \textbf{$1920{\times}1080$} & \textbf{Synth. (phone)} \\
\bottomrule
\end{tabular}
\end{table*}

Table~\ref{tab:deblurring_datasets} situates our dataset in the context of existing deblurring benchmarks. Compared to widely used datasets like GoPro and HIDE, our dataset is significantly larger in both the number of image pairs and scene diversity. With over 42,000 pairs across 843 distinct scenes, and a resolution of $1920{\times}1080$, our dataset enables high-fidelity modeling of motion blur. The use of consumer-grade smartphone footage provides realistic and diverse blur patterns. This combination of scale, diversity, and realism distinguishes our dataset from others, particularly the small-scale, controlled blur datasets like \cite{LevinEval} and \cite{KohlerDataset}, and the relatively lower resolution and limited scenes of older benchmarks. Our contribution fills a critical gap in the deblurring literature by offering a modern, challenging dataset that is designed to better approximate smartphone deployment scenarios. We report results using PSNR and SSIM for consistency with prior work. 
\begin{figure*}
    \centering
    \includegraphics[width=\linewidth]{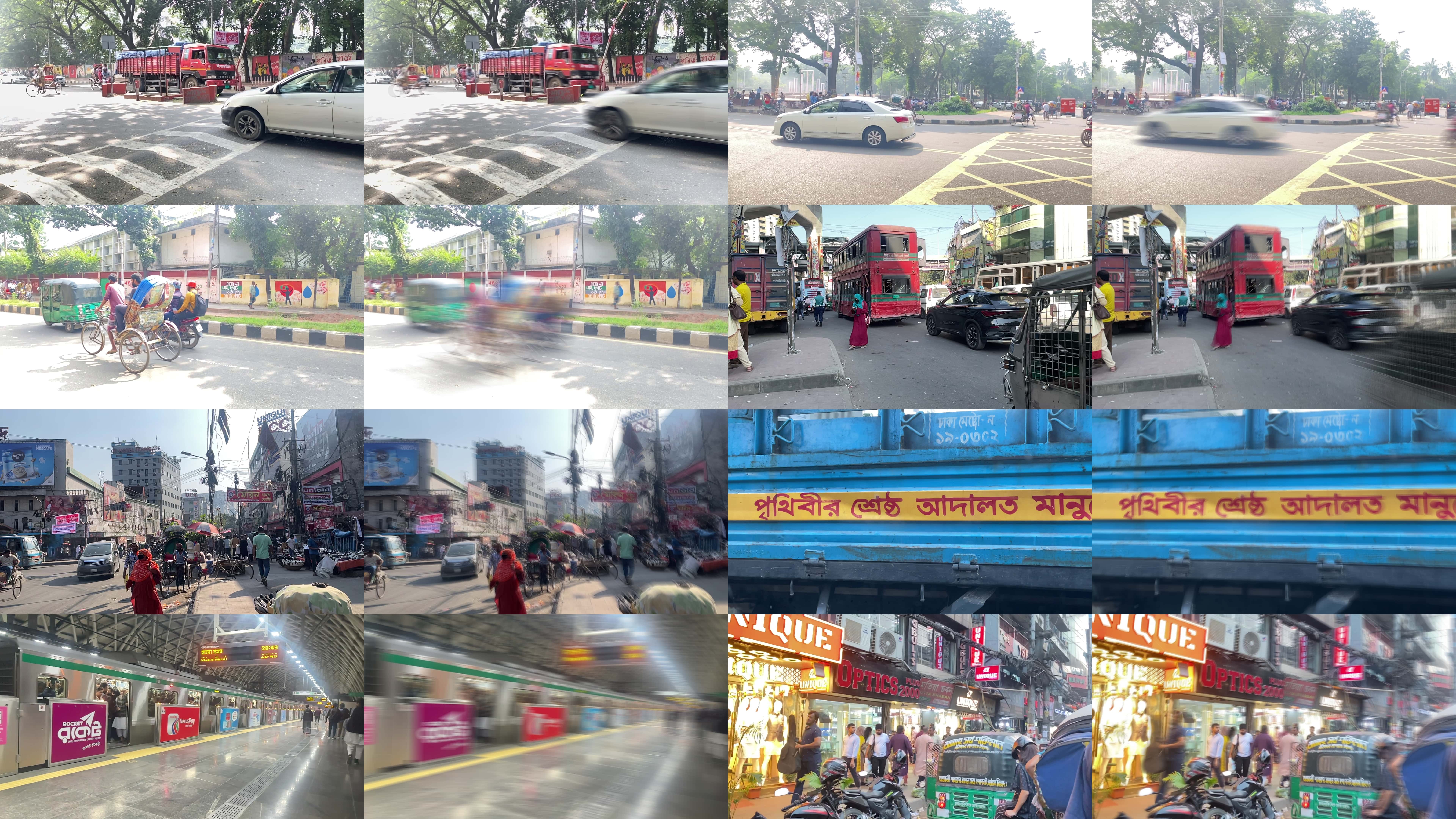}
    \caption{Paired images with sharp and blurred versions.}
    \label{fig:image_grid}
\end{figure*}

\subsection{Visual Examples}
To complement our quantitative evaluation, we present a selection of qualitative examples that demonstrate the diversity and complexity of our dataset. These samples offer a visual understanding of the types of motion blur captured, as well as the realism of the scenes. By showcasing a range of scenarios, including camera shake, fast-moving objects, and human activity, we aim to illustrate the practical challenges faced by current deblurring models.

Figure ~\ref{fig:image_grid} showcases a diverse subset of blurry-sharp image pairs from our dataset, highlighting the variety and realism of the captured scenes. The examples include a wide range of motion types such as object motion, human activity, and camera shake. It also captures a range of lighting conditions and depths of field. These samples illustrate the presence of complex and spatially varying lighting conditions often found in real-world scenarios.

\section{Conclusions}

In this work, we have introduced a novel large-scale image deblurring dataset constructed from high-frame-rate smartphone video, addressing critical limitations of existing benchmarks in scale, realism, and domain relevance. By leveraging the iPhone 15 Pro’s 240 fps slow-motion capture, we synthesized motion blur through temporal averaging and paired each blurred image with a temporally aligned sharp ground truth. This methodology yielded over 42,000 high-resolution blur-sharp pairs encompassing diverse scenes and motion patterns, significantly surpassing the scale of prior datasets.

Our comprehensive benchmarking of state-of-the-art deblurring models, including transformer based architectures and frequency-domain approaches, revealed substantial performance degradation when evaluated on our dataset. This underscores the dataset's complexity and the challenges it poses to existing deblurring techniques, highlighting smartphone-oriented blur and motivating future research toward more robust, generalizable deblurring models.


\subsection{Limitations}

SloMoDeblur is constructed using a single iPhone 15 Pro capture pipeline. This design improves capture consistency, temporal alignment, and reproducibility, but it also means that the dataset reflects the sensor, optics, stabilization, compression, tone mapping, denoising, and sharpening characteristics of this specific smartphone pipeline. Therefore, the benchmark should not be interpreted as covering all smartphone imaging systems. Devices from other manufacturers, including Samsung, Pixel, Xiaomi, and other smartphone families, may exhibit different sensor readout behavior, ISP processing, and residual stabilization artifacts.

The single-device protocol is a deliberate design choice rather than an oversight. Similar to established benchmarks such as GoPro and RealBlur, we prioritize stable imaging characteristics in order to isolate the proposed blur-generation protocol and benchmark scale. Introducing multiple devices at this stage would make it difficult to disentangle blur-model variation from differences in sensor pipelines, optics, and ISP behavior. We therefore view multi-device and cross-ISP capture as an important extension of the present dataset.
\subsection{Future Work}

Building on the strengths of our dataset, several directions can be pursued to enhance its scope and applicability. One natural extension is to augment the dataset with precise motion metadata, enabling the development of motion-aware or physics-informed deblurring models. Another natural direction is to extend SloMoDeblur across multiple smartphone families and ISP pipelines, enabling cross-device evaluation of deblurring models under more heterogeneous mobile imaging conditions. Finally, leveraging our dataset for training generative or diffusion-based models could open new paths for perceptually grounded and temporally consistent deblurring performance. 

Our evaluation relies primarily on PSNR and SSIM, which remain the standard metrics in image restoration benchmarks. While these measures ensure comparability with prior work, they do not fully capture perceptual quality, particularly for generative approaches such as GAN-based models that emphasize realism over pixel-wise accuracy. Metrics like LPIPS or NIQE are better suited to reflect perceptual fidelity, but their inclusion was beyond the scope of this dataset-focused contribution. Our goal in this work was to establish a large-scale, deblurring dataset and to demonstrate its difficulty under widely accepted evaluation protocols. We view the reliance on PSNR/SSIM as a pragmatic choice for benchmarking consistency, while leaving room for future extensions that incorporate perception-oriented metrics for a more comprehensive evaluation of generative methods.

Another important direction is cross-dataset transfer evaluation. A comprehensive comparison between models trained on SloMoDeblur and existing benchmarks such as GoPro or RealBlur requires carefully controlled training protocols and held-out real smartphone test sets. Such evaluation would provide further insight into smartphone-oriented datasets' ability to improve generalization across different imaging conditions.

By providing a dataset that closely mirrors real-world conditions encountered in everyday smartphone photography, we offer a valuable resource for developing and evaluating advanced deblurring algorithms. We anticipate that this dataset will catalyze further research into models that can effectively handle the intricacies of real-world motion blur, ultimately advancing the field of image restoration. Future studies can further validate the proposed blur formation strategy through controlled comparisons with real long-exposure smartphone captures.


\impact{We introduce a new large-scale dataset comprising over 42,000 blurry-sharp image pairs, more than ten times the size of the GoPro dataset, entirely using a modern iPhone. Using an averaging-based approach on high-frame-rate video, we simulate realistic motion blur in unconstrained settings, while the corresponding sharp frames serve as reliable ground truth. This scale, combined with the domain relevance of smartphone imagery, positions our dataset as a new benchmark for image deblurring. It reflects the true challenges of blur encountered in everyday photography and provides a robust foundation for developing models that generalize to the most common real-world use case: photos captured by the average user on a smartphone. While SloMoDeblur primarily supports research in image restoration and computational photography, we acknowledge the potential dual-use nature of image enhancement technologies, including possible surveillance-related misuse. However, the dataset does not introduce a unique surveillance capability beyond existing public deblurring benchmarks and is intended to facilitate responsible research in image restoration. The dataset was captured in public-space settings under the legal and institutional framework applicable at the capture location. Therefore, incidental appearances of pedestrians, vehicles, signage, or other potentially identifiable public-space content may occur. We do not apply face or license-plate blurring, as such post-processing would modify the natural ISP-processed image statistics that the dataset is designed to preserve. The dataset is intended for research use in image restoration and computational photography and is released under the Creative Commons Attribution-ShareAlike 4.0 International (CC BY-SA 4.0) license.}

\acks{This work was supported by the Improving Computer and Software Engineering Tertiary Education Project (ICSETEP), Bangladesh, under Sub-project PIN A-77. The authors also acknowledge the Cognitive Agents and Interaction Lab (CAIL) for providing the necessary research facilities and computational resources.
}



\vskip 0.2in
\bibliography{sample}








\end{document}